# FREQUENCY BASED INDEX ESTIMATING THE SUBCLUSTERS' CONNECTION STRENGTH

LUKAS PASTOREK
Department of Statistics and Probability,
University of Economics, Prague, Faculty of Informatics and Statistics,
W. Churchill sq. 4, 13067 Prague, Czech Republic
email: lukas.pastorek@vse.cz

**Abstract**
*In this paper, a frequency coefficient based on the Sen-Shorrocks-Thon (SST) poverty index notion is proposed. The clustering SST index can be used as the method for determination of the connection between similar neighbor sub-clusters. Consequently, connections can reveal the existence of natural homogeneous. Through estimation of the connection strength, we can also verify information about the estimated number of natural clusters that is a necessary assumption of efficient market segmentation and campaign management and financial decisions. The index can be used as the complementary tool for the U-matrix visualization. The index is tested on an artificial dataset with known parameters and compared with results obtained by the Unified-distance matrix method.*

## 1. Introduction

The Unified-distance matrix method, short U-matrix, (Ultsch and Siemon, 1990) has been widely used framework for visualization of data topology. It was developed as an exploratory tool, which would reveal the spatial distribution and relations preserved through fixed grid topology of Kohonen's self-organizing map (Kohonen, 2001). Kohonen's self-organizing map is the unsupervised neural network, which approximates data topology through iterative learning. Additionally, the self-organizing algorithm incorporates a neighborhood core function, which distributes forces across the neural model grid. A self-organizing neural network projects multidimensional space onto a low dimensional (usually two dimensional) manifold (grid of neurons). A Unified distance matrix, reflects Euclidean distances between neighbor neurons onto a grid through a heat map visualization. On the other hand, a two-dimensional neural grid with a sheet shape induces topology restrictions, which may lead to erroneous conclusions about the multidimensional data topology and the number of clusters. As a correction tool, this paper proposes an application of a frequency based index that may revise and restore input data topology in some cases, when the U-matrix reaches its limitation.

## 2. SST index

A new clustering index, based on the logic of the SST-index is applied to address the issue and limitation of two-dimensional sheet topology.

The Sen-Shorrocks-Thon (STT) index is a sophisticated mathematical concept developed in the 1970s in the study of income and poverty to measure the general income situation of poor people and households (Sen, 1976). The index combined the effects of the relative frequency of the poor with the Gini coefficient and the poverty gap ratio, which measures the average distance between the income of the poor and the poverty threshold. In the following decades, the index underwent modifications (Shorrocks, 1995) to the final form, which enabled its decomposition (Xu and Osberg, 2001). As the real nature of the index is the assignment of weights to each poverty gap (Aguirregabiria, 2006), it can be connected to the theory of clustering and multidimensional space through income replacement by the frequency of projected vectors and the income gap interpreted as a distance from a uniform distribution.

## 3. CSST index

After neural network learning, vector prototypes are distributed across space and a U-matrix of Euclidean distances is constructed. Data distribution between selected pairs of Voronoi regions can be measured with the Clustering SST index (CSST index), which estimates the depth of the gap between selected regions. Before calculation of the index for each pair, input data space assigned to selected Voronoi regions and in-between space is cut, and the frequency distribution is constructed.

Let us draw two Voronoi regions $R_{j1}$ and $R_{j2}$, $j \in \{1,2,\ldots,n\}$. Symbol $n$ denotes the number of Voronoi regions and $j$ is the unique index related to each Voronoi region. For each pair of Voronoi regions, reference vectors $\mathbf{w}_{j1}$ and $\mathbf{w}_{j2}$ respectively, are computed according to the formulas

$$\mathbf{w}_{j1} = \mathbf{c}_{j2} - \mathbf{c}_{j1}, \qquad (1)$$

$$\mathbf{w}_{j2} = \mathbf{c}_{j1} - \mathbf{c}_{j2}, \qquad (2)$$

where $\mathbf{c}_{j1}$ and $\mathbf{c}_{j2}$ are prototype vectors defining Voronoi regions $R_{j1}$ and $R_{j2}$. Each input vector $\mathbf{x}_i, i = 1,\ldots,L$ ($L$-Length of the training dataset), assigned to selected Voronoi regions $R_{jq}, q \in \{1,2\}$; can be defined by its difference from its reference vector $\mathbf{c}_{jq}$

$$\mathbf{v}_{iq} = \mathbf{x}_i - \mathbf{c}_{jq}. \tag{3}$$

Cosine similarities (Deza and Deza, 2006) between $\mathbf{v}_{iq}$ and $\mathbf{w}_{jq}$ are computed according to the formula

$$d_{\cos i} = \frac{\sum_{s=1}^{t}\left(\mathbf{v}_{(s)iq} \times \mathbf{w}_{(s)jq}\right)}{\sqrt{\sum_{s=1}^{t}\left(\mathbf{v}^2_{(s)iq}\right)} \times \sqrt{\sum_{s=1}^{t}\mathbf{w}^2_{(s)jq}}}, \tag{4}$$

where $t$ is the dimension of the input vector space and $s$ is the number attribute in vectors.

Only those vectors $\mathbf{x}_i$ are selected for further analyses, which have positive cosine similarity $(d_{\cos i} > 0)$. Space in-between prototype vectors $\mathbf{c}_{j1}$ and $\mathbf{c}_{j2}$ is consequently cut into imaginary parallel and equal sized sub-regions, and the frequency distribution of the projected input vectors characterizing in-between space is assembled. Each sub-region is described by the frequency of the projected input vectors (satisfying the cosine similarity assumption), which are present in the given sub-region.

A Clustering SST index can be computed according to the following expression

$$CSSTI = \frac{1}{N_D}\left[\sum_{u=1}^{k}[z_u r_u(1+G_{r+})] + \sum_{u=1}^{k}[(1-z_u)r_u(1+G_{r-})]\right], \tag{5}$$

Where $r_u = \left|N_u - \frac{N_D}{k}\right|$. Symbol $N_D$ denotes the number of input vectors, that have $d_{\cos i} > 0$, $k$ is the number of sub-regions, $z_u$ is the Boolean function, which is equal to one, when the $u$-th sub-region's frequency of projected input vectors $\geq \frac{N_D}{k}$. Otherwise, $z_u$ is equal to zero $\left(\forall u, N_u \geq \frac{N_D}{k} : z_u = 1 \wedge \forall u, N_u < \frac{N_D}{k} : z_u = 0\right)$. Symbols $G_{r+}$ and $G_{r-}$ are Gini coefficients, which were computed from the products $z_u r_u$ and $(1-z_u)r_u$, respectively.

## 4. Artificial dataset and experiment

In the testing phase, an artificial dataset was constructed using default Matlab function "*peaks*" (2401 input 3-dimensional vectors). The function creates a data manifold, which includes various rescaled multidimensional Gaussian distributions (1 high peak, 1 deep canyon, 2 medium peaks and 1 medium valley). Peaks are surrounded by flat data surface, see Figure 1 and Figure 2.

Kohonen`s self-organizing map (30x30 grid neurons) and U-matrix were applied using SOM toolbox (Vesanto, 2005) to reveal input data topology, see Figure 3. Bright areas on the U-matrix heat map are supposed to indicate neurons, that are further from their neighbor neurons on the model grid. Dark colored regions reflect the compact cluster of neurons that share the same feature – a very small Euclidean distance to its neighbor grid neurons.

## 5. Results

U-matrix visualization implies the existence of more than two distinct clusters that are divided by the fuzzy and dispersed "mountain range" of the bright color. Unfortunately, visualization does not provide a clear and apparent description of the real topology and an exact determination of the number of clusters is difficult. The cause of this effect is probably the incorrect wrong folding of the Self-organizing map. Further analysis of the proposed a CSSTI index was conducted.

Sixteen distinct neurons were chosen, because of the assumption of the sufficiently large frequency of projected input vectors in each selected region. Other neurons were excluded from the analysis. Spatial relations between selected regions were analyzed using a Euclidean distance matrix and CSSTI matrix, see labels in Figure 1 and Figure 2, and corresponding labeled regions on the U-matrix, Figure 3. Input data vectors were projected to those regions according to the minimum Euclidean distance principle.

According to the U-matrix visualization, regions numbered 5 and 13 and regions 4 and 12 should be distinct regions divided by regions 9 and 8. This fact would have underlined the existence of the at least two opposite distinct clusters.

The Euclidean distance matrix, see Figure 4, indicates that those two regions aren't separated by the great distance. However, simple distance comparisons do not provide us enough strong evidence, as the case of the existence of the only but totally empty sub-region is not excluded. The CSSTI matrix also includes the information about the significance of the sub-regional gap and as we see in Figure 5, regions 5 vs. 13 and 4 vs. 12 show very small values of the CSST index. These facts imply that the U-matrix incorrectly indicates the existence of two opposite distinct regions. In fact, the noted regions can be considered as neighbors. This is a consequence of incorrect map folding, which is apparent in Figure 6.

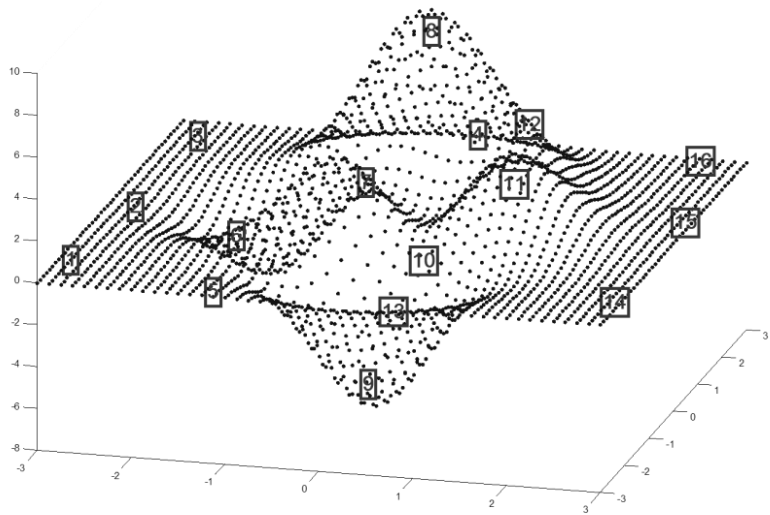

Figure 1 3D Matlab visualization of the Artificial Input Dataset with labeled regions (point of view 1)
Source: Author`s own visualization in Matlab.

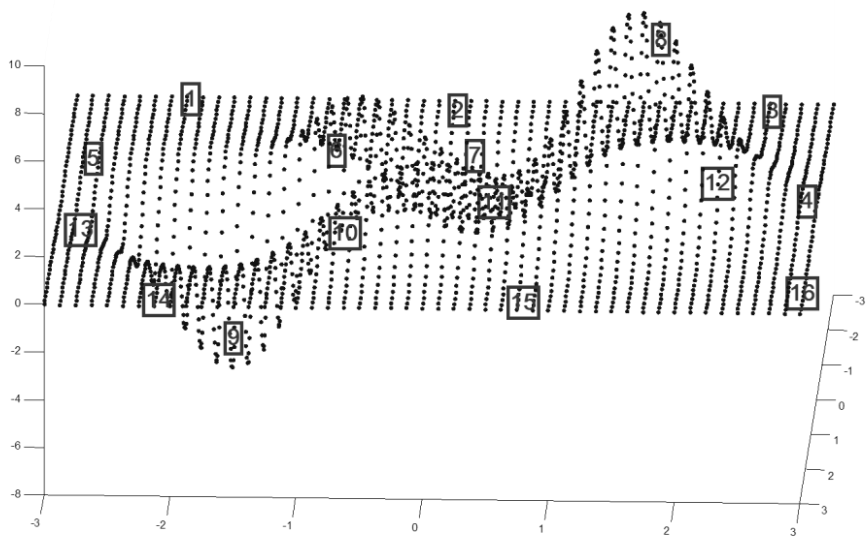

Figure 2 3D Matlab visualization of the Artificial Input Dataset with labeled regions (point of view 2)
Source: Author`s own visualization in Matlab.

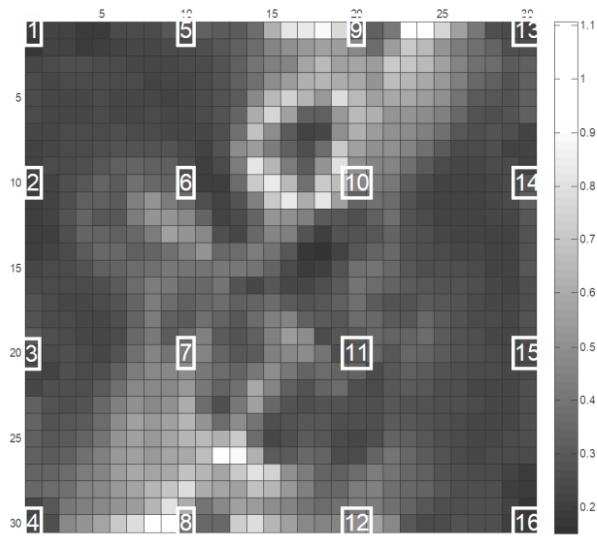

Figure 3 U-matrix visualization with selected labeled regions
Source: Author`s own visualization in Matlab.

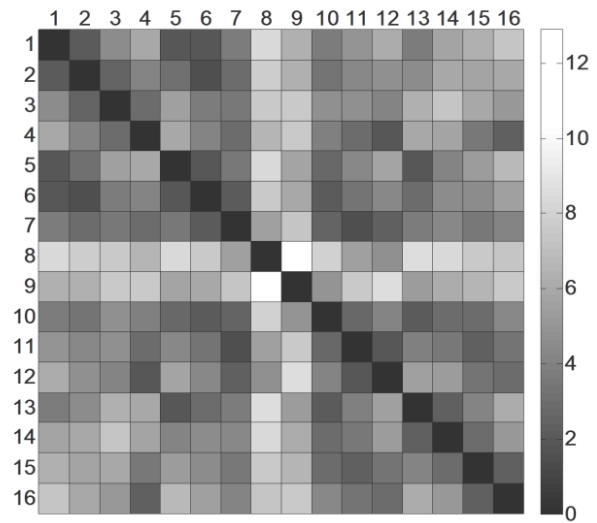

Figure 4 Euclidean distance matrix of selected labeled regions
Source: Author`s own visualization in Matlab.

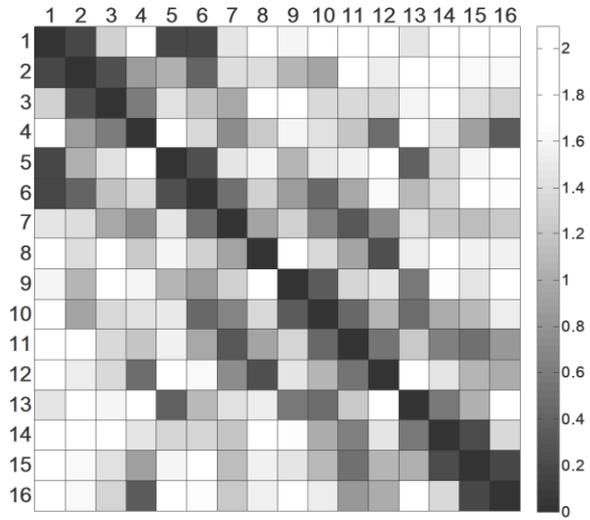

Figure 5 CSST index matrix of selected labeled regions
Source: Author`s own visualization in Matlab.

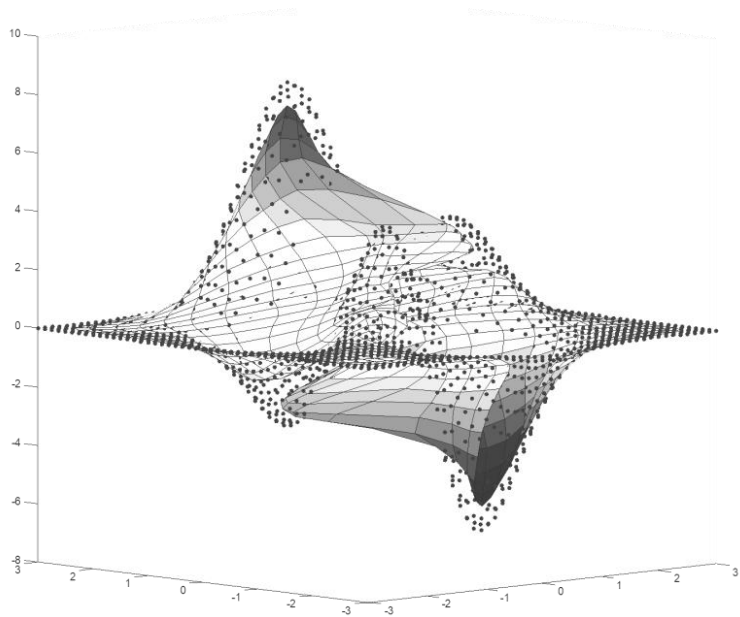

Figure 6 Incorrect folding of Kohonen´s self-organizing map
Source: Author`s own visualization in Matlab.

## 6. Conclusion

The proposed Clustering SST index connects an axiomatic approach to the income distribution of poor individuals with the theory of the density distribution. Moreover, the CSST index can be used as an additional tool for spatial analysis of the input data distribution. In addition, the index was able to distinguish incorrect folding of a self-organizing map. It can also be used independently from the U-matrix to determine the existence of the natural clusters that are split into various model regions. Furthermore, the index could reveal an incorrect localization of the prototype model vectors in the input data space and effect of outliers. On the other hand, the CSST index is not suitable for estimating the depth of the gap between Voronoi sets with low numbers of projected vectors.

**Acknowledgements**
The author acknowledges the use of Matlab SOM toolbox for data analysis and visualization. This work was supported by project IGA VSE F4/17/2013 IG410032.

**References**
1. AGUIRREGABIRIA, V. 2006. Sen-Shorrrocks-Thon index. In: ODEKON, M. (eds.): Encyclopedia of World Poverty. Thousand Oaks, CA: Sage Publications. 2006. ISBN 978-14-129180-7-7.
2. DEZA, M. M. DEZA, E. 2006. Dictionary of distances. (1st ed.). Elsevier, Amsterdam 2006.
3. KOHONEN, T. 2001. Self-organizing maps. (3rd ed.). Springer, Berlin, 2001.
4. PASTOREK, L. VENIT, T. 2012. STT index and its components as the indicators of the monetary poverty and inequality in the Czech and Slovak republic between years 2004 – 2008. In ŽELINSKÝ, T. PAUHOFOVÁ, I. (eds.) Inequality and Poverty in the European Union and Slovakia. Košice: Faculty of Economics TU, 2012. ISBN 978-80-553-1225-5. Pp. 67–74.
5. SEN, A. 1976. Poverty: An ordinal approach to measurement. In: Econometrica. 1976, vol. 44, iss. 2, pp. 587-599.
6. SHORROCKS, A. F. 1995. Revisiting the Sen poverty index. In: Econometrica. 1995, vol. 63, iss. 5, pp. 1225-1230.
7. ULTSCH, A. SIEMON, H. P. 1990. Kohonen's Self Organizing Feature Maps for Exploratory Data Analysis. In WIDROW, B. ANGENIOL, B. (eds.) Proceedings of the International Neural Network Conference (INNC-90), Paris, France, July 9–13, 1990. 1. Dordrecht, Netherlands: Kluwer. ISBN 978-0-7923-0831-7. Pp. 305–308.
8. VESANTO, J. et al. 2005. SOM Tool-box for Matlab 5: Manual and tutorial. Last update 17. 3. 2005, http://www.cis.hut.fi/somtoolbox/ [cit. 20-08-2013]
9. XU, K. OSBERG, L. 2001. How to Decompose Sen-Shorrocks-Thon Poverty Index: A Practitioner's Guide. In: Journal of Income Distribution. 2001, vol. 10, iss. 1-2, pp. 77-94.